\documentclass{article}

\usepackage{arxiv}
\usepackage{algorithm}  
\usepackage[noend]{algpseudocode}  
\usepackage[latin1]{inputenc} 

\usepackage{hyperref}   
\usepackage{url}            
\usepackage{booktabs}       

\usepackage{nicefrac}       
\usepackage{microtype}      
\usepackage{subfigure}
\usepackage{amsmath,amssymb,amsfonts}
\usepackage{graphicx}
\usepackage{textcomp}
\usepackage{xcolor}

\usepackage{booktabs}
\usepackage{mathrsfs}
\usepackage[square,numbers,sort&compress]{natbib}

\DeclareMathOperator{\ODESOLVER}{ODESOLVER}
\DeclareMathOperator{\tuple}{\textit{tuple}}

\title{Neural Ordinary Differential Equation based \\ Recurrent Neural Network
Model}

\author{ \href{https://orcid.org/0000-0001-9051-1370}{\includegraphics[scale=0.06]{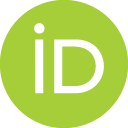}\hspace{1mm}Mansura Habiba} \\
	Cloud Solutions Architect\\
	IBM in Ireland\\
	Dublin, Ireland \\
	\texttt{mansura.habiba@gmail.com} \\
	\And
	\href{https://orcid.org/0000-0003-0521-4553}{\includegraphics[scale=0.06]{orcid.png}\hspace{1mm}Barak A. Pearlmutter} \\
	Department of Computer Science \& Hamilton Institute\\
	Maynooth University\\
	Maynooth, Ireland \\
	\texttt{barak@pearlmutter.net} \\
}

\hypersetup{
pdftitle=Neural Ordinary Differential Equation based \\ Recurrent Neural Network
Model,
pdfsubject={q-bio.NC, q-bio.QM},
pdfauthor={Mansura Habiba, Barak A. Pearlmutter},
pdfkeywords={	Recurrent Neural Network, Time series, Ordinary Differential equation},
}

\begin{document}
\maketitle

\begin{abstract}
	Neural differential equations are a promising new member in the neural network family. They show the potential of differential equations for time series data analysis. In this paper, the strength of the ordinary differential equation (ODE) is explored with a new extension. The main goal of this work is to answer the following questions: (i)~can ODE be used to redefine the existing neural network model? (ii)~can Neural ODEs solve the irregular sampling rate challenge of existing neural network models for a continuous time series, i.e., length and dynamic nature, (iii)~how to reduce the training and evaluation time of existing Neural ODE systems? This work leverages the mathematical foundation of ODEs to redesign traditional RNNs such as Long Short-Term Memory (LSTM) and Gated Recurrent Unit (GRU). The main contribution of this paper is to illustrate the design of two new ODE-based RNN models (GRU-ODE model and LSTM-ODE) which can compute the hidden state and cell state at any point of time using an ODE solver.  These models reduce the computation overhead of hidden state and cell state by a vast amount. The performance evaluation of these two new models for learning continuous time series with irregular sampling rate is then demonstrated. Experiments show that these new ODE based RNN models require less training time than Latent ODEs and conventional Neural ODEs. They can achieve higher accuracy quickly, and the design of the neural network is simpler than, previous neural ODE systems.
\end{abstract}


\label{sec:intro}
Continuous time series is a fundamental data structure for different research areas, i.e., sensor-based IoT, energy consumption, weather, music generation etc. It also has some dynamic characteristics such as high sampling frequency, inconsistent sampling rate, multi-variate, large in length, dynamic and uncertain. Neural network models need to support these characteristics to process continuous time series. Among all the available neural network models, RNN has become the pioneer in terms of time series modelling and processing due to its gating unit and memory storing capacity. The majority of existing RNNs processes continuous time series as a discrete-time sequence with a fixed sampling rate and fixed sampling frequency \cite{neil2016phased}. This makes real-time processing difficult. It also imposes high computation load and memory usage. In addition, due to the dependency on the computation of previous states, such RNN models are prone to some vulnerabilities such as if the time gap between two consecutive observations is too big, it can affect the efficiency of the model adversely. Therefore, such RNNs are only suitable for time series of moderate length with fixed sampling rate, few missing values, and short time intervals between observations.

\cite{chen2018neural, rubanova2019latent} have demonstrated the strength of Neural ODEs \cite{BRYSON62, PEARLMUTTER89A} for processing time series using deep learning.  \cite{chen2018neural} proposes to compute iterative updates of RNN as Euler discretization of continuous transformation \cite{chang2018reversible}. Based on this idea, the continuous hidden dynamics of a neural network can be parametrized by ordinary differential equation \eqref{eq1}. Using the initial value problem to compute $h(t)$ at any time $t$ from the initial hidden units $h(0)$, this continuous-depth model leverages ODE solver ($f$) to define and evaluate models. Latent-ODE \cite{chen2018neural} uses an ODE as an encoder whereas ODE-RNN \cite{rubanova2019latent} uses an RNN as an encoder.

\begin{equation}
\label{eq1}
\frac{d}{dt}h(t) = f(h_{t},t, \delta_{t})
\end{equation}

ODE-RNN \cite{rubanova2019latent} uses encoder and decoder to compute the hidden dynamics of the neural network. The input is fed to an encoder and a different equation solver is used to calculate the hidden state and later the decoder generates the output. This is an auto-regressive model, similar to \cite{chen2018neural}, ODE-RNN can also compute the hidden dynamics at any time $t$. The dynamics between observations is learned rather than pre-defined,  which makes ODE-RNN a suitable model for irregular and sparse data. As the computation is not dependent on the availability of data points, both ODE-RNN and Latent-ODE take more time for training and evaluation. The time depends on the length of the time series and the complexity of data. The focus of this paper is to consider the neural network as a function of continuous time; it therefore does not use any additional encoder or decoder.

In case of time series precessing, RNN based neural network model are already ahead of others. However, different RNN models use a discrete sequence of observations with fixed sampling rate for modelling continuous time series. Therefore, often it is difficult to compute a time series in real time. The main goal of this work is to leverage the continuousness of ODE-NET to design Neural Ordinary Differential Equation based RNN which can reduce the overhead of additional encoding and decoding for simple time series data processing. These proposed models can overcome the limitation of fixed time steps with marginal budget cost in shorter time duration and accurate real time computation.    

This work proposed to design the architecture of different RNN, e.g., GRU \cite{gru} or LSTM \cite{lstm} using an ordinary differential equation, which is referred as ODE-GRU and ODE-LSTM throughout this paper. These proposed models can train the gradients itself within marginal error using the ordinary differential equation solver. All parameters are learned during training. The cell state and dynamics of hidden state are computed using a black-box ordinary differential equation solver. The update and reset functions of existing RNN models are redefined to leverage ODE solver. These proposed models can process continuous time series in real-time with comparatively less computation and memory usage. These models provide a generative process over time series similar to \cite{rubanova2019latent}. We compare proposed models to several RNN variants and find that these new models can perform better when the data is sparse.

\section{Methodology}
\label{sec:proposal}
\begin{figure}[h]
\centering
\includegraphics[width=0.45\textwidth]{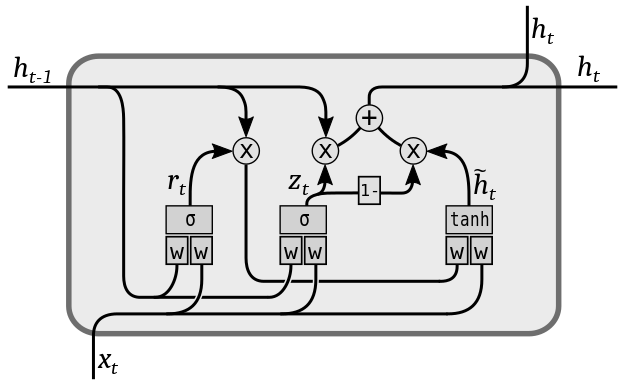}
\caption{Architecture of GRU model}
\label{fig:gru}
\end{figure}

\begin{figure}[h]
\centering
\includegraphics[width=0.45\textwidth]{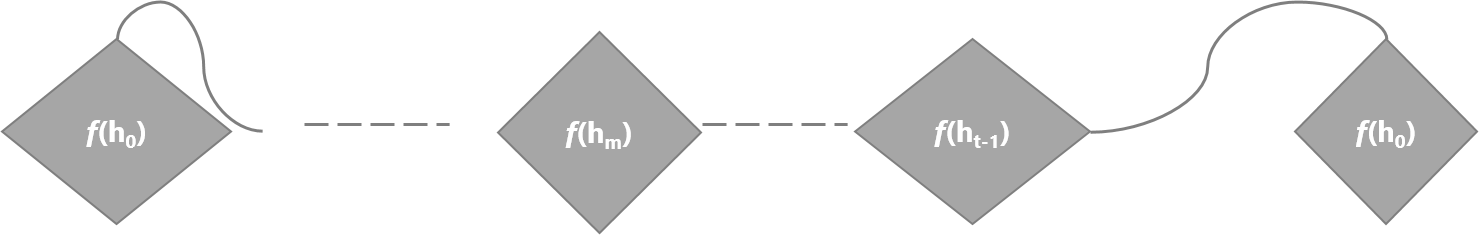}
\caption{Architecture of ODE-GRU model}
\label{fig:ode-gru}
\end{figure}

Fig.~\ref{fig:gru} shows the architecture of the standard GRU model. In a standard RNN cell either GRU or LSTM, the input, output and hidden state $(h_{t})$ are controlled by gating units, i.e reset gate $ (r_{t}) $ and update gate $ (z_{t}) $ as shown by \eqref{eq:rgru} and \eqref{eq:zgru} respectively.   Each cell in standard RNN uses the hidden state of precious cell $(h_{t-1})$  as input along with raw data input.
In this paper, the existing RNN cell is redesigned as a function of ordinary differential equations. Two different RNN models are proposed using the similar cell structure of GRU and LSTM. In this new RNN model, the hidden dynamics is computed using the derivative of hidden state between observations. Fig.~\ref{fig:ode-gru} shows that the proposed model progress over continuous time, where $ h_{0} $ is the initial hidden state passed through an ordinary differential equation (ODE) which uses  update function for solving ODE similar to standard GRU or LSTM cell ($f$). For two different gating units of RNN, two different ODE- based RNN models (i)~ODE-GRU and (ii)~ODE-LSTM are proposed in this paper.

These two proposed models leverage the capabilities of ordinary differential equation solvers. The final cell state $y_{t}$ for the model can be computed using \eqref{eq:odeRNNCell}. \textit{odeRNNCell} refers to an ODE based RNN cell, either GRU-ODE or LSTM-ODE\@.
\begin{subequations}
\label{eq:odeRNNCell}
\begin{align}
\label{eq:yt}
y_{t} &= \ODESOLVER(\textit{odeRNNCell},y_0, t)
\\
\label{eq:yt2}
y_{t} &= \ODESOLVER(\textit{odeRNNCell},\tuple(y_0, h_0), t)
\end{align}
\end{subequations}
$\ODESOLVER$ can be any kind of ode solver, e.g., an Euler or adjoint or implicit method. Models proposed in this paper use a neural ordinary function \textit{odeRNNCell} to compute change or derivative of the hidden dynamics at any time t. \textit{odeRNNCell} is usually an initial value problem function as shown in \eqref{eq:yt} which requires the initial observation $ y_0 $ and hidden state $ h_0 $ at time $t = t_{0}$. In the case of \eqref{eq:yt}, the cell state at time t is computed based on initial time state, and the hidden dynamics is considered as gradient dynamics and computed by the gradients update. \textit{odeRNNCell}, can be either \textit{odeGRU}, as shown in Algorithm~\ref{alg:odegru} or \textit{odeLSTM}, as shown in Algorithm~\ref{alg:updatelstm}. In other case as in  \eqref{eq:yt2}, the cell state or output at any time is computed based on the initial cell state ($y_{0}$) and hidden dynamics ($h_{0}$) at time $t_{0}$. In this implementation, the proposed neural network is initialized with initial hidden state $(h_0)$, and only the initial observation value is passed as the input for the model.

\subsection{Proposed GRU-ODE model}\label{subsec:proposed-gru-ode-model}

\begin{algorithm}
	\caption{ODE-GRU ode solver function}
	\renewcommand{\algorithmicrequire}{\textbf{Input:}}
	\algorithmicrequire: initial hidden and cell state,time series and  weight as well as bias parameters
	\begin{algorithmic}[1]
		\Procedure{odeGRU}{$states,t,parameters$}
		\State$x\gets states [0]$
		\State$h\gets states [1]$
		\State$r_{t}, z_{t}, \tilde{h}_{t}\gets updateGRUFunction(x,h,parameters)$
		\State$h_{t}\gets \tilde{h}_{t}(1-z_{t})$
		\State$o_{t}\gets \sigma(W_{o} h_{t}+b_{o})$
		\State \textbf{return} $\frac{d o_{t}}{d t},\frac{d h_{t}}{d t}$

		\EndProcedure
	\end{algorithmic}
\label{alg:odegru}
\end{algorithm}

Here \emph{states} are the initial hidden state along with the observation value at start time, $t=t_{0}$, t is the time series, \emph{parameters} is a tuple of all weight parameters ($W_{r},U_{r}, W_{r}, U_{z}, W_{h}, U_{h}$) and bias parameters ($b_{r},b_{z},b_{h}$).  \textit{updateGRUFunction} takes the input and compute the traditional GRU update functions:
\begin{subequations}
\begin{align}
\label{eq:rgru}
r_{t}&=\sigma\left(W_{r} x_{t}+U_{r} h_{t-1}+b_{r}\right)
\\
\label{eq:zgru}
z_{t}&=\sigma\left(W_{z} x_{t}+U_{z} h_{t-1}+b_{r}\right)
\\
\label{eq:ch10}
\tilde{h}_{t}&=\tanh \left(W_{r} x_{t}+U_{r} r_{\tau}+b_{r}\right)
\\
\label{eq:ogru}
o_{t}&= \sigma\left(W_{o} h_{t}+b_{o}\right)
\end{align}
\end{subequations}
ODE-GRU pass the changes to the ODE solver and uses only the change in hidden state as \eqref{eq:dh_dt}. The proposed RNN models do not carry  hidden state of precious cell $(h_{t-1})$ as like standard RNN models. Only the derivative of hidden state over time$ (frac{d\left(h_{t}\right)}{d t} )$ is progressed as a function of time . 
\begin{subequations}
\begin{align}
\label{eq:dh_dt}
\frac{d h_{t}}{d t} &= \frac{d \tilde{h}_{t}}{d t}(1-z_{t})
\end{align}
\end{subequations}

Algorithm~\ref{alg:traingru} describes the complete flow for the training the model using the ODE-GRU model we propose.
\begin{algorithm}
	\caption{Train the ODE-GRU model}
	\renewcommand{\algorithmicrequire}{\textbf{Input:}}
	\algorithmicrequire: Initial condition ($y_{0}$ and $h_{0}$), continuous time series(t) and Initial parameter
	\begin{algorithmic}[1]
		\Procedure{train}{$y_{0}$, $h_{0}$, target, t, initalParams}
		\State$input\gets tuple(y_{0}, h_{0})$
		\State $trainedParams$ $\gets$ optimizer (gradient(GRADF),
				\newline \hspace*{5em}$initalParams$,  $iterations$,  
				\newline \hspace*{5em}$\text{callback} = \text{GRADFUNCTION.backward}())$
		\State \textbf{return} $trainedParams$
		\EndProcedure
	\end{algorithmic}

\begin{algorithmic}[2]
	\Procedure{GRADF}{$input,t,initalParams, target$}
	\State $prediction$ $\gets$ ODESOLVE(
        $odeGRU$, $input$, $t$, \newline \hspace*{14em}$initalParams$)
	\State $loss\gets lossFunction(prediction, target)$
	\State \textbf{return} $loss$
	\EndProcedure

\end{algorithmic}
\label{alg:traingru}
\end{algorithm}

Here any \textit{optimizer} and \textit{lossFunction} can be used based on the complexity of the dataset. \textit{GRADF} is the gradient function uses by optimizer.

\subsection{Proposed ODE-LSTM Model}\label{subsec:proposed-gru-lstm-model}
 Proposed ODE-LSTM model predicts the output ($y_{t}$) at any time t using \textit{odeLSTM} ODE solver using the initial value of the observation ($y_{0}$) at time $t=t_{0}$. Therefore, instead of computing each state based on its previous state individually, the predicted output is generated as a function of time.~\eqref{eq:31} describes ODE Solvers, those have been used in this work:

\begin{equation}
\label{eq:31}
y_{t} = \ODESOLVER(\textit{odeLSTM},\tuple(y_0, h_0, c_0), t)
\end{equation}

Here, \textit{odeLSTM} is the function that constructs the dynamics of hidden as well as cell state and call the ODE solver to compute all the gradients at once. The algorithm of \textit{odeLSTM} is discussed in \ref{alg:updatelstm}. This model uses same input parameter as ODE-GRU, the initial observation value, $ (y_{0}) $ and the initial hidden state of the network $h_{0}$. Besides, this model also uses the initial cell state $(c_{0})$ as an input parameter. This proposed ODE solver function construct the network as an initial value problem function, therefore providing the initial condition is important.

\begin{algorithm}
	\caption{ODE-LSTM update function}
	\renewcommand{\algorithmicrequire}{\textbf{Input:}}
	\algorithmicrequire: initial hidden and cell state,time series and  weight as well as bias parameters
	\begin{algorithmic}[1]

		\Procedure{odeLSTM}{$states,t,param$}
		\State$x\gets states [0]$
		\State$h\gets states [1]$
		\State$c\gets states [2]$
		\State$r_{t}, z_{t},c_{t}, h_{t}, o_{t} \gets updateLSTMGate(x,h,c,param)$
		\State$cell \gets \frac{d f_{t}}{dt} * c + \frac{d i_{t}}{dt} * c_{t}$
		\State$hidden \gets \frac{d o_{t}}{dt} *\frac{d \textit{cell}}{dt}$
		\State \textbf{return} $\frac{d o_{t}}{dt},\textit{hidden},\textit{cell}$

		\EndProcedure
	\end{algorithmic}
\label{alg:updatelstm}
\end{algorithm}

Here \textit{states} contains the initial hidden state, initial cell state and the observation value at start time, $t=t_{0}$, $t$ is the time series, $\textit{param}$ is a tuple of all weight parameters $(W_{r},U_{r}, W_{r}, U_{z}, W_{h}, U_{h})$ and bias parameters $(b_{r},b_{z},b_{h})$. $\textit{updateLSTMGate}$ takes the input and compute the traditional LSTM update functions:
\begin{subequations}
\begin{align}
\label{eq:ilstm}
i_{t}&=\sigma_{i}\left(x_{t} W_{x i}+h_{t-1} W_{h i}+w_{c i} \odot c_{t-1}+b_{i}\right)
\\
\label{eq:flstm}
f_{t}&=\sigma_{f}\left(x_{t} W_{x f}+h_{t-1} W_{h f}+w_{c f} \odot c_{t-1}+b_{f}\right)
\\
\label{eq:clstm}
c_{t}&=f_{t} \odot c_{t-1}+i_{t} \odot \sigma_{c}\left(x_{t} W_{x c}+h_{t-1} W_{h c}+b_{c}\right)
\\
\label{eq:olstm}
o_{t}&=\sigma_{o}\left(x_{t} W_{x o}+h_{t-1} W_{h o}+w_{c o} \odot c_{t}+b_{o}\right)
\\
\label{eq:hlstm}
h_{t}&=o_{t} \odot \sigma_{h}\left(c_{t}\right)
\end{align}
\end{subequations}

Different update gate function such as input ($i_{t}$), forget($f_{t}$) and output($o_{t}$) use the typical sigmoidal nonlinearities $\sigma_{i},\sigma_{f}, \sigma_{o}$ and cell ($c_{t}$) and hidden ($h_{t}$) activation functions uses tanh nonlinearities. $\sigma_{c}$ and $\sigma_{h}$ along with weight parameters( $W_{h i}$, $W_{h f}$, $W_{h o}$, $W_{x i}$, $W_{x f}$, $W_{x o}$), connect the different inputs and gates with the memory cells and outputs, as well as biases ($b_{i}$, $b_{f}$, and $b_{o}$). Both biases and connection weights ($w_{c i}$, $w_{c f}$, and $w_{c o}$) are optional for further influence the update of input, forget, and output gates.

ODE-LSTM pass the changes in input, forget, hidden gate to the ODE solver and uses only the change in smaller fraction to compute the final output, hidden and cell state using \eqref{eq:output}, \eqref{eq:hidden} and \eqref{eq:cell}:

\begin{subequations}
\begin{align}
\label{eq:cell}
c_{t}&= \frac{d f_{t}}{dt} * c_{0} + \frac{d i_{t}}{dt} * c_{t}
\\
\label{eq:output}
\tilde{o}_{t}&= \frac{d o_{t}}{dt}
\\
\label{eq:hidden}
h_{t} &= \tilde{o}_{t} * \frac{d c_{t}}{dt}
\end{align}
\end{subequations}

Algorithm~\ref{alg:trainolstm} describes the complete flow for the training the model using proposed ODE-LSTM model. \textit{odeLSTM} is computing all gradients at once for the model. The parameters are learned during training using the loss function. The change in hidden data dynamics is computed and learned during training session of the proposed model. Similar to algorithm~\ref{alg:traingru}, optimizer uses gradient function (\textit{GRADF})  

\begin{algorithm}[htpb]
	\caption{Train the ODE-LSTM model}
	\renewcommand{\algorithmicrequire}{\textbf{Input:}}
	\algorithmicrequire: Initial conditions ($y_{0}$, $h_{0}$), input$(t)$, $initialParams$
	\begin{algorithmic}[1]
		\Procedure{train}{$y_{0}$, $h_{0}$, $c_{0}$, $target$, $t$, $initalParams$}
		\State$\textit{input} \gets (y_{0}, h_{0})$
		\State $trainedParam$ $\gets$ optimizer(gradient($GRADF$),\newline
				\hspace*{5em} $initalParams$, $iterations$, \newline
				\hspace*{5em} $\text{callback} = \text{GRADFUNCTION.backward}())$
		\State \textbf{return} $trainedParameters$
		\EndProcedure
	\end{algorithmic}

	\begin{algorithmic}[2]
		\Procedure{GRADF}{$input,t,initalParams, target$}
		\State$prediction\gets ODESOLVE(odeLSTM, input, t,$\newline \hspace*{14em}$initalParams)$
		\State$loss\gets lossFunction(prediction, target)$
		\State \textbf{return} $loss$
		\EndProcedure
	\end{algorithmic}
\label{alg:trainolstm}
\end{algorithm}
\section{Results}
A number of experiments are done in order to understand the dynamic characteristics of the dataset.These experiments also show how the complexity of the data influence the proposed models. 

\subsection{Curve datasets}
\begin{figure*}[ht]
	\includegraphics[width=0.48\textwidth]{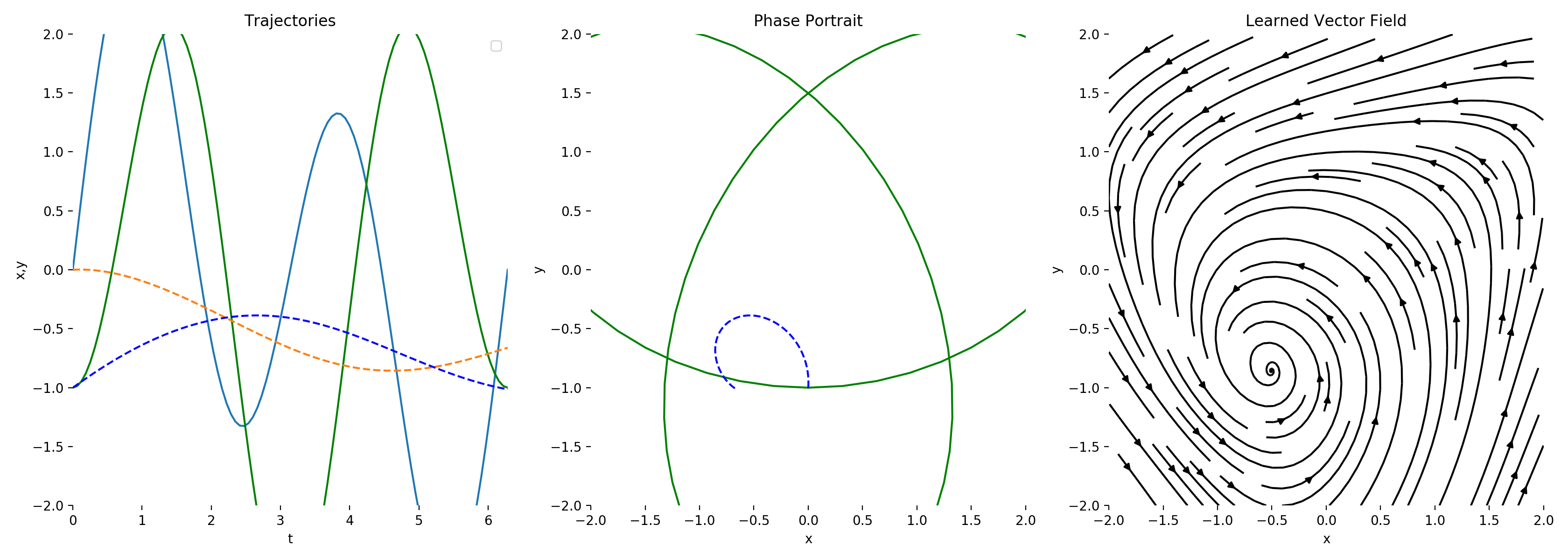}
	\hfill
	\includegraphics[trim=13 10 12 14,width=0.48\textwidth]{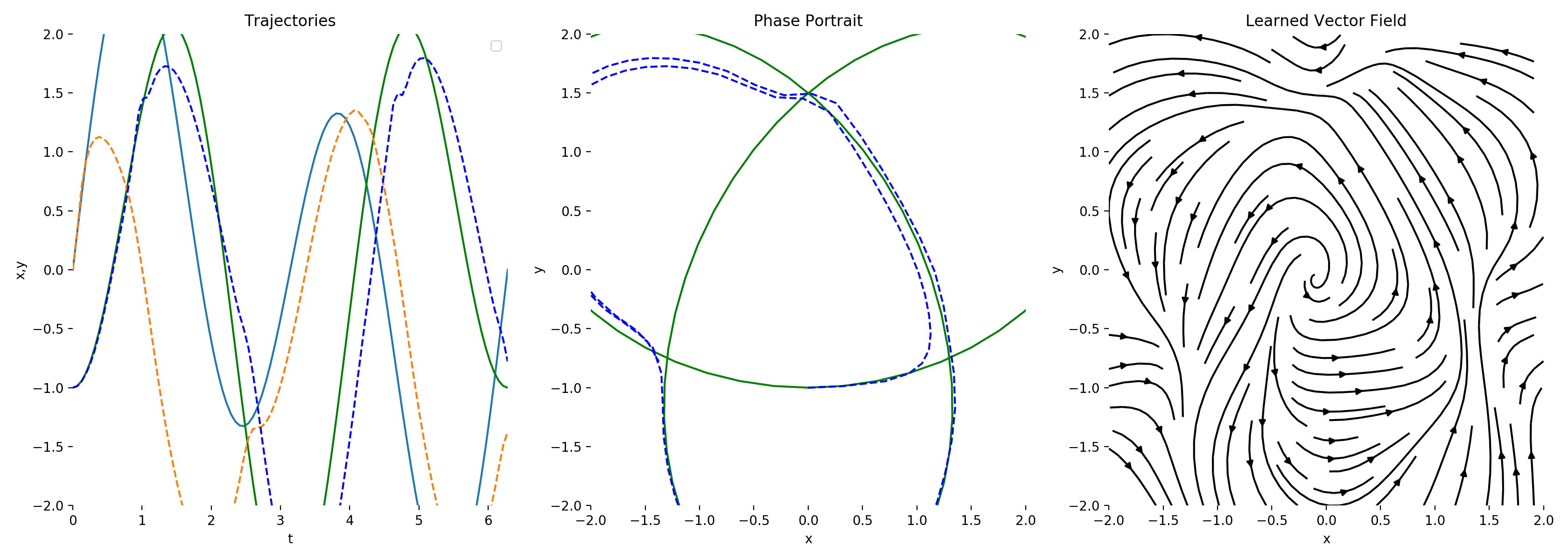}
	\caption{Learning Tri-Knot curve function at iteration 1(Left) and 20(Right) }
	\label{fig:tk}
\end{figure*}

In order to understand the learning mechanism of proposed models, two different curve time series, e.g., (i)~eight-curve with low complexity, (ii)~tri-knot with medium complexity  are chosen. Each of them are of different complexity. This experiment helps to understand the impact of time series flow as well as the complexity and sparseness of time series dataset on the learning of the proposed model. The learning vector for tri-knot curve also demonstrate that the proposed model can learn itself comparatively faster than traditional LSTM and other ODE based RNN \cite{chen2018neural, rubanova2019latent}. Fig.~\ref{fig:tk} shows that the proposed ODE-GRU model needs only 20 iterations to identify learning vector and the curve dynamics.

\begin{figure*}[ht]
	\centering
	\includegraphics[width=0.45\textwidth]{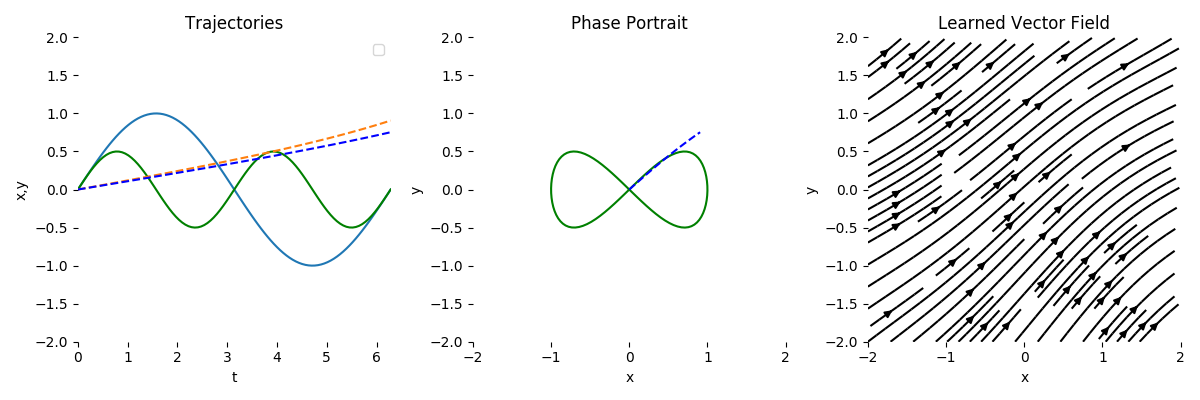}
	\hfill
	\includegraphics[trim=13 10 12 14,width=0.48\textwidth]{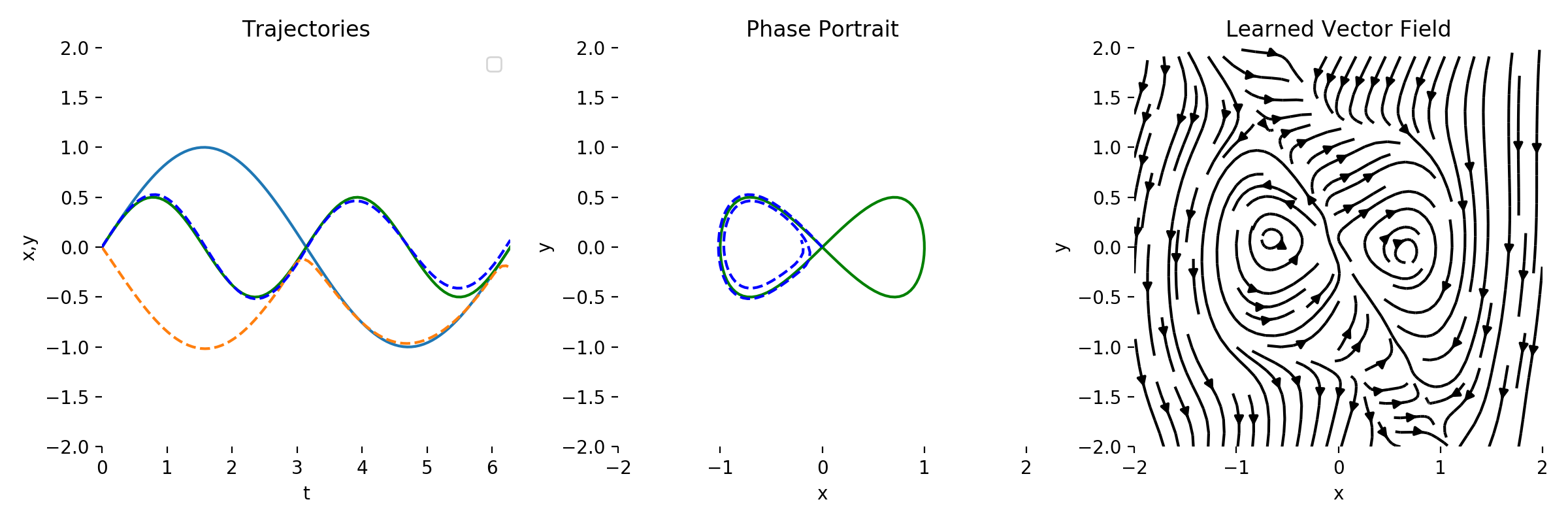}
	\caption{Learning eight curve function at iteration 1 (Left) and 20(Right) }
	\label{fig:ec20}
\end{figure*}

Fig.~\ref{fig:ec20} and shows the learning vector of 1st and 20th iteration for eight curves. This demonstrates that the learning growth is very fast, even for only 50 hidden dimensions. This proposed model can play a significant role to identify the types of dynamic of the corresponding time series data. These experiment shows that with few parameters these proposed models can learn time series data faster than traditional RNN cells. 

For further comparative performance evaluation, two different datasets (a Toy dataset and a Human activity dataset) are used.

\subsection{Toy dataset}
\label{subsec:toy-dataset}
A simple time series  with variable sampling rate or frequency is used to learn the dynamics of proposed model by mimicking a spiral ODE.  The spiral ODE used in this test consists of 1000 spiral  and each spiral has 100 irregularly sampled time points.
In \cite{implemnetation}, a simple Feed Forward Neural network (FFN) is used for ODE solver as shown in~\eqref{eq:yt}.

\begin{algorithm}
def forward(self, t, y):\\
\hspace*{2em} return self.net(y**3)
\end{algorithm}

\begin{equation}
\label{eq:pred_y}
\textit{pred}_y = \textit{odeint}(\textit{FFN}, \textit{batch}_{y_0}, \textit{batch}_t)
\end{equation}

\begin{figure*}[http]
\centering
\includegraphics[width=0.48\textwidth]{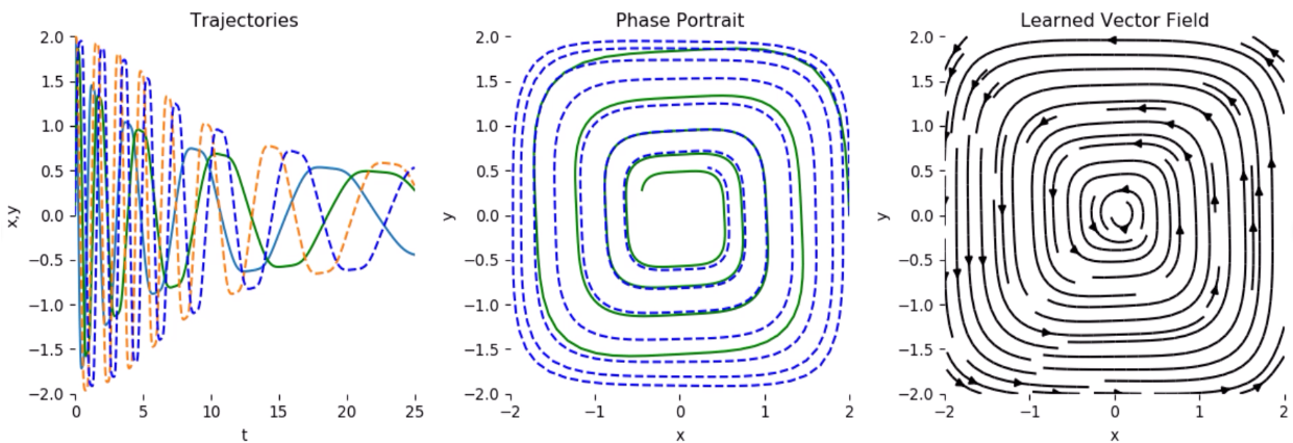}
\hfill
\includegraphics*[trim=13 10 12 14,width=0.48\textwidth]{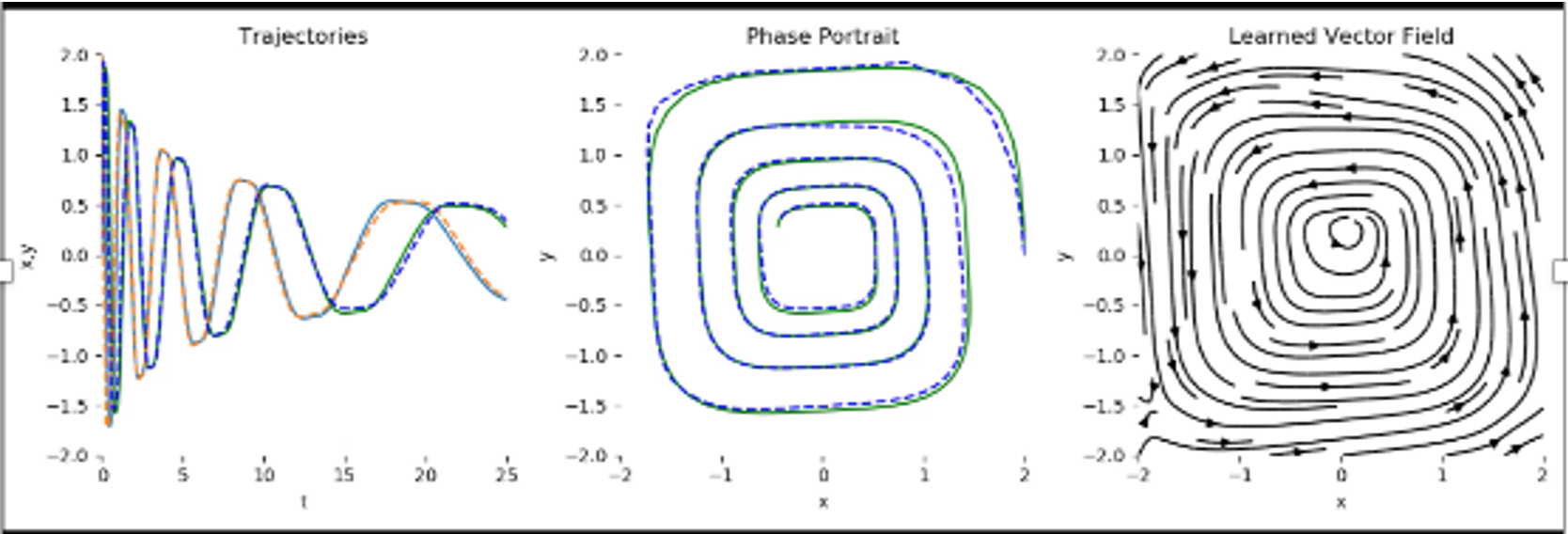}
\caption{Result of ODE neural network (left) and the proposed ODE-GRU (right) after 25 iterations of training.}
\label{fig:ode-25}
\end{figure*}

Fig.~\ref{fig:ode-25} shows the result after the 25 iterations of the implementation of \cite{chen2018neural} (left) and ODE-GRU (right). This demonstrate that the proposed ODE-GRU can reduce the training time significantly, which is one of the limitations of  existing work \cite{chen2018neural}.

\subsection{Human Activity dataset}
\begin{figure*}[http]
\hfill
\subfigure[Latent-ODE]{\includegraphics[width=0.45\textwidth]{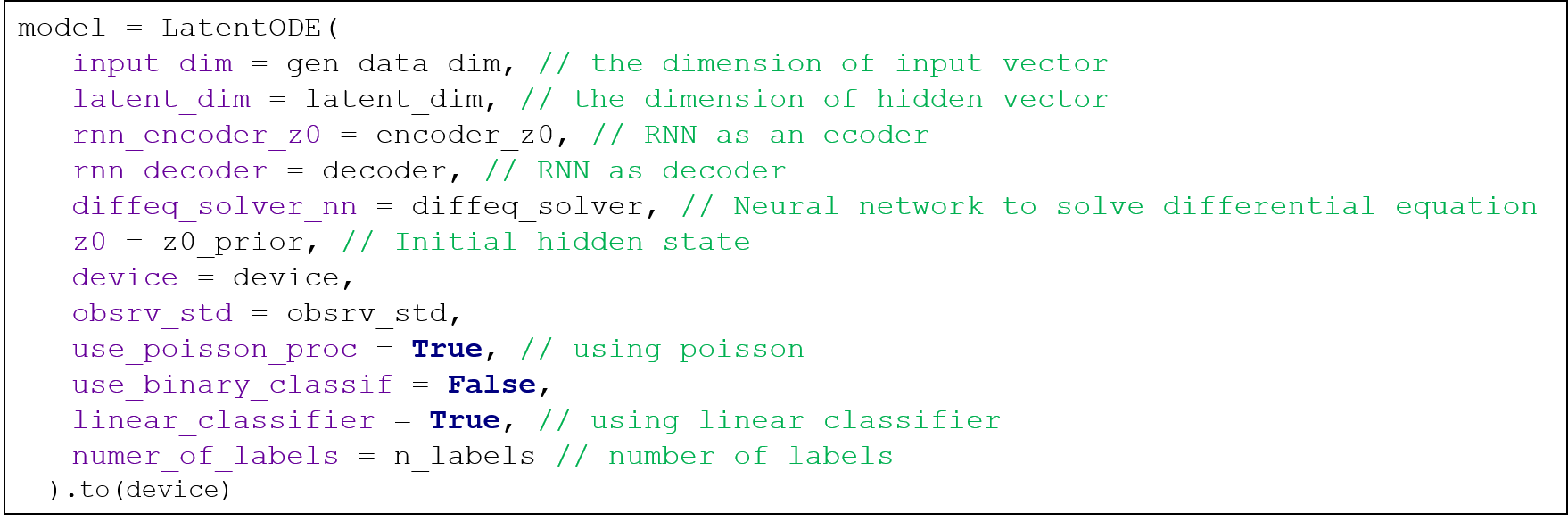}}
\hfill
\subfigure[Proposed ODE-GRU/ODE-LSTM]{\includegraphics[width=0.5\textwidth]{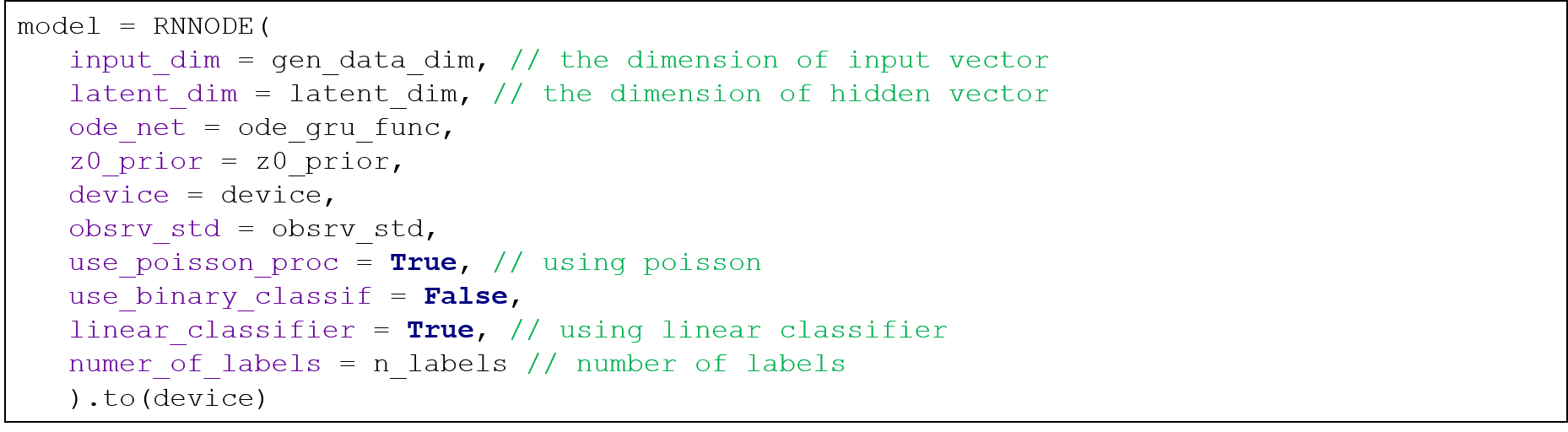}}
\caption{Configuration for neural network used in Human activity experiment}
\label{fig:latent-ode}
\end{figure*}

\begin{figure}[http]
\centering
\includegraphics[width=0.85\textwidth]{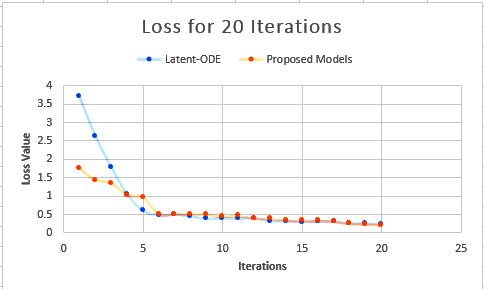}
\caption{Loss value for 20 iteration for Latent-ODE and proposed models}
\label{fig:loss}
\end{figure}

For this experiment and a comparative performance evaluation of proposed model and \cite{rubanova2019latent}, the human activity recognition dataset \cite{anguita2013public} is used. This dataset consists of 6 different activities along with 12 feature. After normalization this dataset can be transferred to a time series of 6554 sequence at 211 time points. The configuration of Latent-ODE \cite{rubanova2019latent} for this comparative analysis is shown at Fig.~\ref{fig:latent-ode}a. There are 3 different neural networks are used for  Latent-ODE, i.e., encoder, differential equation solver and decoder. On the other hand, the proposed neural network only requires a single neural network in order to compute the derivative of hidden dynamics as shown in Fig.~\ref{fig:latent-ode}b. Proposed models need one a single neural network which is used by a simple differential equation solver function as shown in Algorithm~\ref{alg:odegru}. The proposed models are very simple in structure in comparison to other neural ODE, therefore , the training time is less than \cite{rubanova2019latent} and \cite{chen2018neural}. We have also compare with standard GRU and LSTM. On important characteristics of proposed model is the difference in loss between consecutive iteration of the training process is comparatively lower than other neural network models. The reason is it only computes the derivative of hidden dynamic $\frac{d}{dt} h_{t}$ at any time $t$, therefore the proposed models progress through iterations with small changes in loss value, as shown in Fig.~\ref{fig:loss}.

The significant contributions of these proposed models are as following 
\begin{enumerate}
\item \textit{Simple architecture:} The architecture design of proposed models is straightforward, it uses the update functions of standard RNN models and ODE differential equation solver to design RNN models as a function of the continuous time. As RNN models are already pioneering in terms of time series modelling, these proposed models leverage the strength of RNN and neural ODE in a single neural network.
\item \textit{Data dynamics:} Due to continuous and straightforward characteristics, these models are very suitable for understanding the data dynamics. These models can compare the dynamics of different dynamics to understand the data. This feature can significantly contribute to design data-driven system. These models can also participate in explainable machine learning. 
\item \textit{Faster Training:} As these proposed models learn the change in hidden dynamics of the data along with the gating unit of RNN, it takes comparatively less time for training and evaluation without compromising the accuracy of the result. 
\item \textit{Fewer parameters:} In comparison to standard RNN cells and different neural ODEs, even fewer weight, as well as bias parameters, can achieve the result even faster for proposed models. 
\item \textit{Memory Usage:} Proposed models use the autograd to compute graphs for storing previous state. However, not all previous states are relevant as it only focuses on the parameters of the model and the hidden dynamics. Based on the complexity of the dataset, the graph length can vary. Data with higher complexity uses more memory than a simple dataset. But a successful implementation with properly cleaning unnecessary state can control the memory usage for complex dataset. 
\end{enumerate}

\section{Conclusion}
Unlike traditional RNN models, these proposed models do not need to discretize continuous time series into discrete-time sequences. In contrast to generic GRU model, this proposed ODE-GRU model can update the dynamics at irregularly sampled time points. Besides, the proposed ODE based two RNN models leverage the distinctive architecture of GRU and LSTM gating unit to model data-driven , event-driven asynchronously sampled input of continuous series. These ODE-GRU and ODE-LSTM models have significant advantages. Any large sequence can be learned within a short time and fewer parameters. ODE-GRU computes continuously, therefore it is updated at every time slot where there is a change $\frac{d}{dt} h_{t} >0$. If there is no change in consecutive time slot, ODE solver would compute $\frac{d}{dt}h_{t}=0$. As a result, no additional masking vector is required to distinguish missing values. The significant contribution of these proposed models is they are effortless to compute with faster training as well as evaluation time.

\bibliographystyle{unsrt}
\bibliography{references}  

\end{document}